\documentclass[a4paper,11pt,onecolumn]{article}
\usepackage[margin=25mm,footskip=18pt]{geometry}
\usepackage{graphicx}
\usepackage[authoryear,round]{natbib}
\usepackage{amsmath,amssymb}
\usepackage{xcolor}
\usepackage{booktabs}
\usepackage{times}
\usepackage{microtype}
\usepackage{url}
\definecolor{revisionblue}{RGB}{0,65,155}
\definecolor{queryred}{RGB}{165,25,25}
\newif\ifshowchanges
\showchangestrue 

\newcommand{\R}{\mathbb{R}}

\graphicspath{{figures/}}

\title{DS3M: Deep structured state space models for nonlinear system identification on modern hardware%
  \mbox{%
    \thanks{This work was funded by the Deutsche Forschungsgemeinschaft
    (DFG, German Research Foundation) -- Project-ID 531152215 -- CRC 1701.}\kern0.45em%
    \thanks{The authors would like to thank Karim Cherifi for his valuable comments.}\kern0.45em%
    \thanks{This work has been submitted to IFAC for possible publication.}\kern0.35em%
  }}
\author{Marco Roschkowski \qquad Hannes Gernandt\\[0.5em]
  \small University of Wuppertal, Wuppertal, Germany\\
  \small (e-mail: \{roschkowski,gernandt\}@uni-wuppertal.de).}
\date{}

\begin{document}
\maketitle
\begin{abstract}
Neural state-space identification learns dynamical models from measured inputs and outputs. SUBNET combines this task with an encoder that estimates initial states from short measurement histories. Its continuous-time version requires many successive neural-network evaluations during numerical integration, resulting in slow training speed which is a major limitation for practical use cases. We propose deep structured state space models (DS3M), which combine compact nonlinear dynamical components in parallel and in series. Custom GPU kernels implemented with Triton reduce the launch and memory overhead of their sequential evaluation. We explain this construction, describe a port-Hamiltonian variant, and present various benchmark comparisons of accuracy and training time.
\end{abstract}

\section{Introduction}
System identification constructs a dynamical model from measured inputs and
outputs. Such a model describes how a system responds to new inputs and
provides a basis for simulation, monitoring and controller design
\citep{schoukens2019roadmap}. This is useful when a model derived from physical
laws is unavailable, too expensive to evaluate, or contains unknown
constitutive relations. For example, friction, saturation and other
nonlinear effects may be difficult to describe accurately from first
principles but are visible in measured trajectories.

We consider control systems of the typical form
\begin{equation}\label{eq:dynamics}
  \begin{split}
    \dot x(t)&=f(x(t),u(t)),\\
    y(t)&=g(x(t),u(t)).
  \end{split}
\end{equation}
Here the input $u(t)\in\mathbb{R}^m$ and output are measured, whereas the state $x(t)\in\mathbb{R}^n$ is generally
unobserved. The identification problem is to determine a model of
\eqref{eq:dynamics} from sampled input--output data. Besides classical identification approaches \cite{nelles2020nonlinear}, neural networks provide flexible parameterizations of the unknown functions $f$ and $g$
in \eqref{eq:dynamics}. However, learning a state equation also requires
initial states for the simulated trajectories. SUBNET addresses this
problem by learning an encoder from a short measurement history to a model
state \citep{beintema2021encoder,beintema2023subnet}. The same encoder and
state equation are trained on many short subsequences. This approach was
extended to continuous-time models in \citet{beintema2023ctsubnet} and to
port-Hamiltonian models in \citet{moradi2026oephnn}. Recently, a variant to obtain symbolic representations of port-Hamiltonian systems was presented in \cite{messaoudi2026ph}.

Accurate models can nevertheless be expensive to train. Each trajectory
requires successive state updates, and a numerical integration step can
itself require several evaluations of a neural network. Small state
equations therefore generate many small, dependent computations. A graphics
processing unit (GPU) can process different trajectories in parallel, but
the repeated overhead to move data between different memory locations in the GPU can significantly slow down the computation.

\begin{figure}[h]
    \centering
    \includegraphics[width=0.58\linewidth]{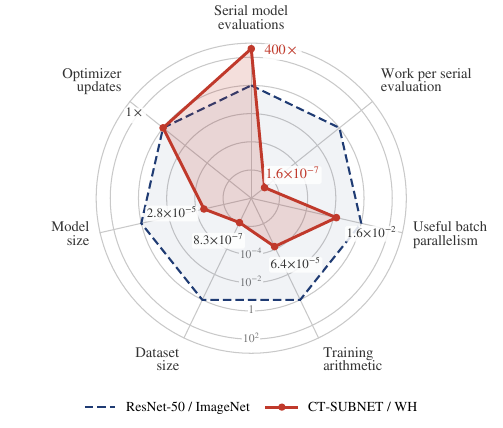}
    \caption{A comparison of relevant training parameters for the standard machine learning task of training ResNet50 \cite{he2016resnet} on Imagenet \cite{russakovsky2015imagenet} and SUBNET \cite{beintema2023subnet} on the Wiener-Hammerstein benchmark from \cite{schoukens2009wiener, schoukens2026benchmarking}. }
    \label{fig:workload}
\end{figure}

Our starting point is the comparison of training workloads between training CT-SUBNET on the Wiener-Hammerstein dataset and training ResNet50 on Imagenet, see \cite{he2016resnet, russakovsky2015imagenet}.
This comparison is depicted in Figure~\ref{fig:workload}.
We can see that the latter task, a standard machine learning task, requires significantly fewer serial model evaluations and benefits from substantially higher batch parallelism while requiring a higher training arithmetic. In particular, training ResNet50 on Imagenet benefits significantly more from modern hardware accelerators such as GPUs. The most important reason for this are the sequential model evaluations during the rollout of CT-SUBNET which often requires hundreds of serial evaluations for a single training step. The present work aims to speed up the training time of CT-SUBNET on modern accelerator hardware. To do this, we implemented custom GPU kernels using Triton \cite{tillet2019triton}. This allows to drastically reduce the number of kernel launches and thus accelerate the training speed on GPUs by a significant margin. However, since this forces the dynamics to be executed in a single program instance, it requires that the weights have a relatively small size to fit into the relevant SRAM and that the neural network architecture is known in advance.
Although this suffices to match state of the art results on standard benchmark datasets \cite{schoukens2026benchmarking} and in some cases even allows us to supersede them by additional hyperparameter tuning, this represents a limitation for future applications on identifying larger systems. Although we do not analyze the neural scaling laws of our method as in \cite{roschkowski2026neural}, our method does in principle allow to be scaled to arbitrary model capacities while already having favorable properties on conventional benchmark datasets as shown below.

Our approach combines two choices. First, custom GPU kernels fuse the
repeated evaluation of compact state equations and their integration steps.
Each of the blocks uses a deep neural network to encode the dynamics, hence the name deep structured state space models (DS3M). The second choice is to connect several of these blocks in a serial interconnection.

\section{Proposed method}

In this section, we describe the interconnection of CT-SUBNET components, its
implementation with Triton, and the port-Hamiltonian parametrisation.

\begin{figure}[h]
  \centering
  \includegraphics[width=\textwidth]{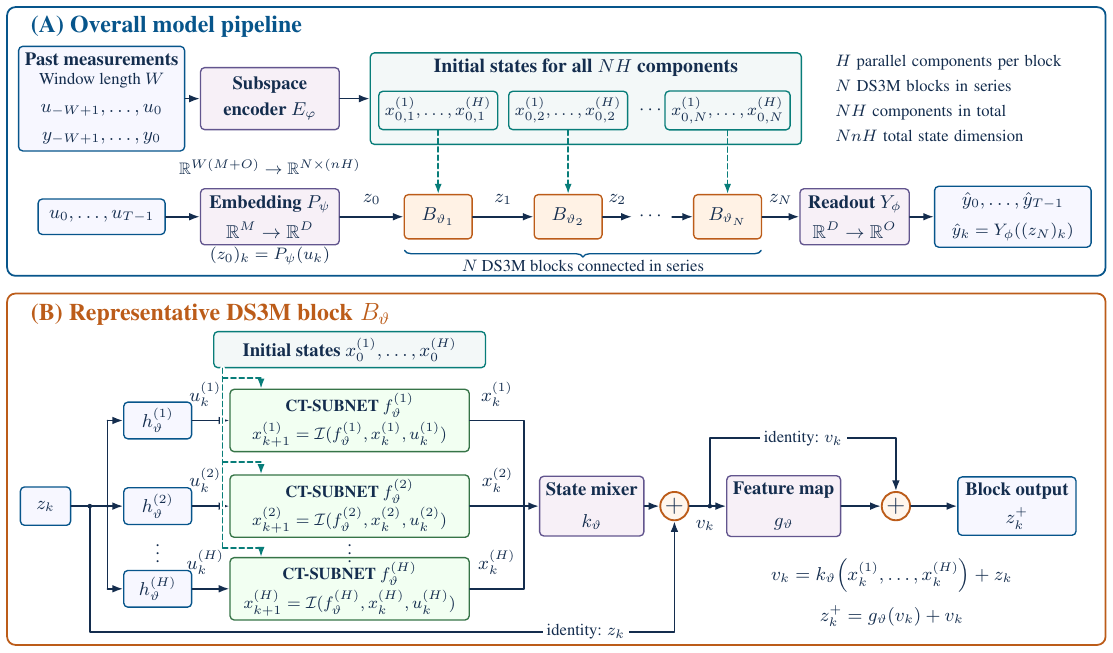}
  \caption{DS3M architecture. Panel (A) shows the overall model pipeline. Here the encoder supplies initial states for
  all $NH$ components. Each block propagates $H$ CT-SUBNETS in parallel,
  mixes their states, and adds the result to its input features.
  The $N$ blocks are connected in series.}
  \label{fig:architecture}
\end{figure}

\subsection{Recap of SUBNET and CT-SUBNET}
First, we recall how SUBNET learns initial states and dynamics jointly, and where
the sequential computational cost originates.

In the output-error setting considered here, the measured output is modelled
as the output of a deterministic state equation plus measurement noise.
For each training subsequence, an encoder receives a finite history of
inputs and outputs and estimates the state at the start of a simulation.
The estimated state is then propagated using the recorded inputs, and the
simulated outputs are compared with the measurements. The encoder, state
equation and output map are optimised jointly. This avoids measuring the
state or treating the initial state of every subsequence as an independent
optimisation variable.

The typically nonlinear encoder approximates a state reconstruction map. It compresses the
information about the past that is needed to predict subsequent behaviour.
The term \emph{subspace encoder} connects this construction to state
reconstruction in subspace identification, see \cite{verhaegen1992subspace, vanoverschee1994n4sid}.

Training uses short, possibly overlapping, subsequences sampled from the
available records. Their simulation errors are averaged over a batch.
Different subsequences can be evaluated in parallel, and truncation reduces
the length of the simulations through which derivatives must be
propagated. Thus, SUBNET already provides parallelism across training
windows. Within each window, however, the next state still depends on the
current state.

In CT-SUBNET, a neural network parametrises the state derivative, and a
numerical integrator produces the sampled state transition
\citep{beintema2023ctsubnet}. For sampling interval $\Delta t>0$, we write
\begin{equation}\label{eq:integrate}
    x_{k+1}=\mathcal I(f_\vartheta,x_k,u_k),
 \qquad \dot x=\tau^{-1}f_\vartheta(x,u),
\end{equation}
where $\tau>0$ scales the state derivative. The map $\mathcal I$ integrates
over one sampling interval and discretizes the differential equation on the right in \eqref{eq:integrate}. We refer to \cite{linares2026controlled} for a more detailed discussion of discretization methods.
Under a zero-order hold, $u_k$ is kept constant
on that interval. For example, explicit Euler gives
$x_{k+1}=x_k+(\Delta t/\tau)f_\vartheta(x_k,u_k)$.
The ratio $\Delta t/\tau$ therefore affects the size of the state update
and the numerical conditioning of training. For more general explicit methods with $s$ stages we require $s$ vector-field evaluations
per integration step. With one step per sample, advancing a window of
$T$ samples takes approximately $sT$ such evaluations.

Kernel fusion combines several operations, such as neural network evaluations and numerical integration steps, into a single GPU kernel. This reduces the overhead associated with repeatedly launching kernels and transferring intermediate results between GPU memory locations. The state updates remain sequential, since the computation of $x_{k+1}$ depends on $x_k$.
We will now describe how we use kernel fusion to obtain an architecture which is able to utilize the high level of parallelization of modern GPUs better.

\subsection{DS3M blocks}

Our starting point is to use a single kernel implemented in Triton to speed up the SUBNET rollout. Here we use an implementation which keeps the weights inside the registers throughout the entire kernel execution, but other designs are also possible. Since the registers can only store a fixed amount of parameters, the size of the SUBNET parameters is limited by this approach. This issue limits future applications to larger systems. To address it, we propose to execute the forward of $H$ SUBNETs in parallel in a single block, paralleling the multi-head construction in transformers, \cite{vaswani2017attention}. Secondly, we use a serial interconnection of $N$ such blocks with appropriate MLP blocks that mix the dynamics of each component.

We first define one DS3M block as shown in Figure~\ref{fig:architecture} and then explain how the encoder and readout complete the model.

Let $z=(z_0,\ldots,z_{T-1})$ be a sequence of features with
$z_k\in\R^D$. A block contains $H$ dynamical components, each with state
$x_k^{(j)}\in\R^n$ and internal input $u_k^{(j)}\in\R^m$. Its neural
networks have dimensions
\[
\begin{aligned}
h_\vartheta^{(j)} &: \R^D\to\R^m,&
f_\vartheta^{(j)} &: \R^{n+m}\to\R^n,\\
k_\vartheta &: \R^{nH}\to\R^D,&
g_\vartheta &: \R^D\to\R^D,
\end{aligned}
\]
where $\vartheta$ collects the trainable weights of each DS3M block and $j=1,\ldots,H$. Its operations are
\begin{align}
u_k^{(j)}&=h_\vartheta^{(j)}(z_k),\nonumber\\
x_{k+1}^{(j)}&=\mathcal I(f_\vartheta^{(j)},x_k^{(j)},u_k^{(j)}),
\label{eq:block}\\
v_k &= k_\vartheta\!\left(
x_k^{(1)},\ldots,x_k^{(H)}\right) + z_k \nonumber \\
z_k^+&=
g_\vartheta( v_k) +  v_k.\nonumber
\end{align}
The map $h_\vartheta^{(j)}$ selects a learned input for each CT-SUBNETS. Their states are then combined by $k_\vartheta$ and $g_\vartheta$.
We use skip connections in the last two equations of \eqref{eq:block}, see \cite{he2016resnet}.

We write $z^+=B_\vartheta(z,x_0)$,
$x_0=(x_0^{(1)},\ldots,x_0^{(H)})\in\R^{nH}$ and
$z,z^+\in\R^{T\times D}$. The full model uses $N$ blocks of the form \eqref{eq:block}.
The measured inputs satisfy $u_k\in\R^M$ and the outputs satisfy
$y_k\in\R^O$. An embedding $P_\psi:\R^M\to\R^D$ produces the initial
features, and a readout $Y_\phi:\R^D\to\R^O$ produces the predictions.
The subspace encoder
$E_\varphi:\R^{W(M+O)}\to\R^{N\times nH}$ receives
$(u_{-W+1},\ldots,u_0,y_{-W+1},\ldots,y_0)$.
Writing its $i$th output row as $x_{0,i}^{\top}$, the complete model is
\[
\begin{aligned}
 (z_0)_k&=P_\psi(u_k),\\
 z_i&=B_{\vartheta_i}(z_{i-1},x_{0,i}),
 \qquad i=1,\ldots,N,\\
 \hat y_k&=Y_\phi((z_N)_k).
\end{aligned}
\]
The resulting latent state dimension is $NnH$. The loss function for one subsequence is the mean squared error
\[
  \mathcal L(\hat y,y)
  =\frac1{TO}\sum_{k=0}^{T}\|\hat y_k-y_k\|^2,
\]
where $\|\cdot\|$ is the Euclidean norm.  Training averages this loss over the sampled subsequences and updates all
maps that are shown in Fig.~\ref{fig:architecture} jointly.

\subsection{Efficient evaluation with Triton}

We explain how the numerical recurrence can be evaluated with fewer GPU launches and memory transfers.

A GPU kernel is a function executed on the graphics processor. One kernel launch starts a collection of program instances, each performing the specified operations on part of the data, typically on different batch instances. Triton is a programming language
and compiler for writing such functions using operations on blocks of
arrays \citep{tillet2019triton}. In a conventional PyTorch implementation, matrix products, activation functions and integration stages can result in many small kernel launches at every sample. Their launch and memory overheads can become significant when the state equation is small.

Our custom kernels combine evaluations of the neural state equation and
the integration steps. A time loop can execute inside a program instance,
while independent training windows and the $H$ components supply parallel
work. Intermediate quantities can be reused in registers or other fast
on-chip memory. The benefit comes from reducing dispatch overhead and
unnecessary transfers to the GPU's main memory.

For the chosen fused implementation, efficiency depends on the size of the
data that must remain available during the computation. Large state
equations increase register and shared-memory use and may reduce the
number of program instances that run concurrently. This motivates compact
components. There is a major alternative design option using the more general CUDA language \cite{nickolls2008scalable}. This amounts to implementing a larger kernel in which different program instances are allowed to produce parts of the data and to communicate with each other. This design has, for example, already been used in persistent RNNs \cite{diamos2016persistent}.

\begin{figure}[h]
  \centering
  \includegraphics[width=0.82\linewidth]{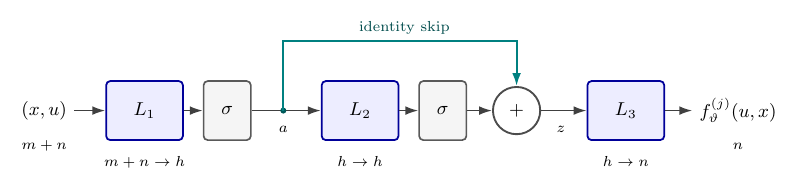}
  \caption{Architecture for MLP that are used to parametrize $f_\theta^{(j)}$ in the DS3M blocks.}
  \label{fig:mlp}
\end{figure}

We use a residual multilayer perceptron (MLP) layout for the component
dynamics, as illustrated in Fig.~\ref{fig:mlp}. A realisation with the state
and internal input concatenated into $v=\left(\begin{smallmatrix}
x\\u    
\end{smallmatrix}\right)\in\R^{n+m}$ has the form
\[
 a=\sigma(L_1v),\qquad
 f_\vartheta^{(j)}(x,u)
   =L_3\bigl(a+\sigma(L_2a)\bigr).
\]
Here $L_1,L_2,L_3$ are trainable affine maps of dimensions
$n+m\to h$, $h\to h$ and $h\to n$, respectively, and the activation function $\sigma$ acts
componentwise. The addition of $a$ to $\sigma(L_2a)$, often called skip connection, is in the hidden layer which typically has the most parameters and tends to stabilize training.

Stacking dynamical layers is also used in structured state space models
\citep{gu2022s4,gu2024mamba}, including applications to dynamical systems
\citep{hu2026dynamicalssm}. Mamba employs updates of the form
$x_{k+1}=A(u_k)x_k+B(u_k)$, with coefficients available from the inputs.
Closedness of affine-linear transformations under composition are used to substantially improve the computational efficiency.
DS3M instead retains nonlinear dependence on the state
and reduces the cost of sequential integration through fusion. Since the dynamics of DS3M are significantly more expressive, this allows to identify systems with significantly fewer states than used conventional SSMs with linear dynamics in $x$.

\subsection{Port-Hamiltonian dynamics}
As an alternative to the standard DS3M block we also can impose a physics-based structure on each component to incorporate
dissipation into the learned state equation~\cite{ISOPHNN,moradi2026oephnn}. This is done by assuming a port-Hamiltonian structure for $f_\vartheta^{(j)}$ variant of our method. In this case,
\[
\begin{aligned}
 f_\vartheta^{(j)}(x,u)
 ={}&\bigl(J_\vartheta^{(j)}(x)-R_\vartheta^{(j)}(x)\bigr)
       \nabla \mathcal{H}_\vartheta^{(j)}(x)+B_\vartheta^{(j)}(x)u,
\end{aligned}
\]
where $J_\vartheta^{(j)}(x),R_\vartheta^{(j)}(x)\in\R^{n\times n}$,
$B_\vartheta^{(j)}(x)\in\R^{n\times m}$ and
$\mathcal{H}_\vartheta^{(j)}:\R^n\to[0,\infty)$. 
Here $J_\vartheta^{(j)}(x)$ is skew-symmetric and
$R_\vartheta^{(j)}(x)$ is positive semidefinite.
For a compact kernel we use
\[
 \mathcal{H}_\vartheta^{(j)}(x)=\frac{1}{2}\|x\|^2,\qquad
 R_\vartheta^{(j)}(x)
   =\operatorname{diag}(r_{\vartheta,1}^{(j)}(x),\ldots,
                         r_{\vartheta,n}^{(j)}(x)),
\]
where softplus, $\sigma(a)=\log(1+\mathrm e^a)\geq0$, enforces nonnegative
diagonal entries in the output layer of $r_{\vartheta, i}^{(j)}$. The restriction $\mathcal{H}_\vartheta^{(j)}(x) = \frac{1}{2}\|x\|^2$ substantially simplifies the Triton kernel implementation since no backward through a neural network is needed in the forward implementation. On the other hand, the choice that $R^{(j)}_\vartheta$ is diagonal instead of parameterizing it via a Cholesky factorization as in \cite{ISOPHNN} as another simplification which we choose to further stabilize training.
Suppressing indices, the continuous-time component satisfies
\[
 \tau\frac{\mathrm d}{\mathrm dt}\mathcal{H}(x)
 =-\nabla \mathcal{H}(x)^\top R(x)\nabla \mathcal{H}(x)
    +u^\top B(x)^\top\nabla \mathcal{H}(x).
\]
which implies that for zero input, i.e.\ $u=0$ the energy $t\mapsto \mathcal{H}(x(t))$ is nonincreasing.

\section{Comparison with benchmark results}

\begin{table}[h]
\centering
\caption{Runtime comparison of PyTorch baselines and the
fused Triton kernel of a single CT-SUBNET forward with $n=6, m=3, h=32$, batch size $128$, $\sigma = \mathrm{SiLU}$ and $\mathrm{RK}4$ integrator.
Timings include forward and backward passes. Here 'compliled' means that the torch.compile functionality is used.
}
\label{tab:triton_speedup}

\small
\setlength{\tabcolsep}{3pt}
\begin{tabular}{@{}clrrr@{}}
\toprule
$T$ & \textbf{Baseline}
    & \textbf{Base [ms]}
    & \textbf{Triton [ms]}
    & \textbf{Speedup} \\
\midrule
100 & CPU eager     & 154.651 & 1.850 & $83.61\times$ \\
100 & CPU compiled  &  52.854 & 1.850 & $28.58\times$ \\
100 & CUDA eager    & 280.963 & 1.850 & $151.91\times$ \\
100 & CUDA compiled &  79.145 & 1.850 & $42.79\times$ \\
\midrule
200 & CPU eager     & 370.954 & 3.779 & $98.16\times$ \\
200 & CPU compiled  & 117.881 & 3.779 & $31.19\times$ \\
200 & CUDA eager    & 603.999 & 3.779 & $159.82\times$ \\
200 & CUDA compiled & 153.942 & 3.779 & $40.73\times$ \\
\bottomrule
\end{tabular}
\end{table}

In this section, we explain the outcomes of our experiments.
Each experiment is conducted on an Nvida RTX 3090 card with PyTorch $2.11.0+\mathrm{cu}126$ and Triton $3.6.0$. We use $\mathrm{float}32$ as the numerical accuracy to implement the kernels.
We compare the reported simulation errors and training times on some standard nonlinear
system identification benchmarks \cite{schoukens2026benchmarking}.

Throughout the experiments, we keep the following parameters constant. Each $k_\vartheta$, $h_\theta^{(j)}$ and $P_\psi$ consists of a single linear layer. Moreover, $g_\vartheta, Y_\phi$ are always MLPs with two hidden layers and $64$ hidden units. Outside of $f_\vartheta^{(j)}$, we always use the SiLU activation function, see \cite{elfwing2018sigmoid}. The embedding dimension is $D=16$ and each layer has only one head which means $H=1$. For the subspace encoder, we use an MLP with two hidden layers and hidden dimension $64$.
We follow the standard practice of standard-scaling the inputs and outputs before training. The Adam optimizer, see \cite{kingma2014adam}, with initial learning rate $0.001$, weight decay $0$, gradient clipping $0.001$ which is applied separately to the model and encoder parameter groups and cosine annealing learning rate schedule, see \cite{loshchilov2016sgdr}. A random seed is used for reproducibility.
We evaluate the models at a given frequency, keep an exponential moving average (EMA) with warm start after $2000$ steps and decay $0.999$. We apply validation-based checkpoint selection and choose the model with the best validation score, including EMA and non-EMA snapshots.
The parameters that are varied for different experiments are shown in Table~\ref{tab:parameters}.

For each block, the weight matrices $W_1$, $W_2$, and $W_3$ of $f_\vartheta^{(j)}$ are initialized using Kaiming-uniform initialization (with $a=\sqrt{5}$) and subsequently scaled by factors of $0.5$, $0.5$, and $0.1$, respectively. The corresponding biases are initialized uniformly in $[-1/\sqrt{d_{\mathrm{in}}},\,1/\sqrt{d_{\mathrm{in}}}]$, where $d_{\mathrm{in}}$ denotes the input dimension of the respective layer, and are not rescaled. All other parameters are initialized using the PyTorch default initializations.

\begin{table}[h]
\centering
\caption{Parameters that change throughout the experiments.}
\label{tab:parameters}
\small\setlength{\tabcolsep}{3pt}
\resizebox{\textwidth}{!}{%
\begin{tabular}{@{}lcccccccccccc@{}}
\toprule
\textbf{Dataset} & \textbf{N} & \textbf{n} & \textbf{m} & \textbf{h}
& $\sigma$ & $\mathcal{I}$ & $T$
& training iterations & $\Delta t/\tau$
& batch size & validation frequency & history length\\
\midrule
Silverbox
& $2$ & $2$ & $2$ & $32$
& $\mathrm{SiLU}$ & RK4 & $150$
& $200,000$ & $0.5$
& $128$ & $500$ & $50$\\
EMPS
& $1$ & $3$ & $2$ & $64$
& $\mathrm{tanh}$ & Heun & $2500$
& $15,000$  & $0.25$
& $256$ & $50$ & $20$\\
WH
& $2$ & $4$ & $4$ & $32$
& $\mathrm{SiLU}$ & RK4 & $200$
& $500,000$  & $1$
& $128$ & $500$ & $50$\\
WH (port-Hamiltonian)
& $2$ & $4$ & $4$ & $32$
& $\mathrm{SiLU}$ & RK4 & $200$
& $200,000$  & $0.3$
& $128$ & $500$ & $50$\\
ParWH
& $2$ & $8$ & $8$ & $32$
& $\mathrm{SiLU}$ & RK4 & $250$
& $800,000$  & $0.5$
& $128$ & $500$ & $50$\\
\bottomrule
\end{tabular}%
}
\end{table}

The comparison of the RMSEs and the training times for the considered methods are shown in Tables~\ref{tab:wh_comparison}--\ref{tab:silverbox_comparison}. SUBNET and CT-SUBNET provide direct comparisons with the underlying encoder approach. Additionally we consider methods with the best scores in the benchmark, which are polynomial nonlinear state-space models (PNLSS), polynomial
nonlinear autoregressive models with exogenous inputs (PNARX), dynoNet,
and nonlinear linear-fractional-representation models (NL-LFR)
\citep{paduart2012pnlss,forgione2021dynonet,floren2025inference,
schoukens2026benchmarking}. We also include the training times in the table which are numbers reported in the literature on non-identical hardware.

\begin{table}[h]
\centering
\caption{Wiener--Hammerstein benchmark.}
\label{tab:wh_comparison}
\small\setlength{\tabcolsep}{5pt}
\begin{tabular}{@{}lcc@{}}
\toprule
\textbf{Method}&\textbf{Training time}&\textbf{RMSE [mV]}\\
\midrule
SUBNET&$>200\,\mathrm h$&$0.24$\\
PNLSS&n.a.&$0.42$\\
\textbf{DS3M}&$1.5\,\mathrm h$&$0.24$\\
\textbf{DS3M (port-Hamiltonian)}&$2.4\,\mathrm h$&$0.28$\\
\bottomrule
\end{tabular}
\end{table}

\begin{table}[h]
\centering
\caption{EMPS benchmark. The dynoNet record uses a different historical data split.}

\label{tab:emps_comparison}
\small\setlength{\tabcolsep}{3pt}
\begin{tabular}{@{}lcc@{}}
\toprule
\textbf{Method}&\textbf{Training time}&\textbf{RMSE [mm]}\\
\midrule
CT-SUBNET&$1$--$24\,\mathrm h$&$4.6$\\
dynoNet (original)&n.a.&$2.6$\\
PNARX&$1$--$60\,\mathrm{min}$&$2.6$\\
\textbf{DS3M}&$0.5\,\mathrm h$&$1.8$\\
\bottomrule
\end{tabular}
\end{table}

\begin{table}[h]
\centering
\caption{Parallel Wiener-Hammerstein benchmark.}.
\label{tab:five_comparison}
\small\setlength{\tabcolsep}{2.5pt}
\begin{tabular}{@{}lcc@{}}
\toprule
\textbf{Method}&\textbf{Training time}&\textbf{RMSE}\\
\midrule
NL-LFR&$1$--$60\,\mathrm{min}$&$[0.20,0.40,0.41,0.47,0.83]$\\
\textbf{DS3M} &$2.7\,\mathrm h$&$[0.20,0.38,0.26,0.30,0.49]$\\
\bottomrule
\end{tabular}
\end{table}

\begin{table}[h]
\centering
\caption{Silverbox}
\label{tab:silverbox_comparison}
\small\setlength{\tabcolsep}{4pt}
\begin{tabular}{@{}lcc@{}}
\toprule
\textbf{Method}&\textbf{Training time}&\textbf{RMSE [mV]}\\
\midrule
NL-LFR&$1$--$60\,\mathrm s$&$[0.29,0.33,0.26]$\\
SUBNET&$1$--$4\,\mathrm d$&$[0.36,1.4,0.32]$\\
\textbf{DS3M}&$0.5\,\mathrm h$&$[0.23,0.74,0.23]$\\
\bottomrule
\end{tabular}
\end{table}

In Table~\ref{tab:wh_comparison}, we can see that the proposed method reaches the test error of $0.24 \,\mathrm{mV}$ with significantly less training time than the SUBNET baseline on the Wiener-Hammerstein benchmark. Here we also consider the port-Hamiltonian variant which yields slightly worse results with $200,000$ instead of $500,000$ training steps which we have chosen for use since its training steps take longer.
Moreover, we can see in Table~\ref{tab:emps_comparison} that the proposed method is able to reach a significantly better performance on the EMPS benchmark. We used the higher training sequence length of $T=2500$ This became practically possible because of the training time speedup we achieved. 
Table~\ref{tab:five_comparison} shows that our method is able to achieve favorable performance on the parallel Wiener-Hammerstein benchmark.
In Table~\ref{tab:silverbox_comparison} we can see that our method allows to decrease the training time and improve the performance relative to the SUBNET baseline on the Silverbox benchmark. This also allows to close a significant part of the gap to methods that perform better on this benchmark such as NL-LFR. However, we still see that our method tends to perform well on the first and third test trajectory but less so on the second test trajectory which leaves the training distribution more. This phenomenon is discussed in more detail in \cite{beintema2021encoder} and our method does not seem to resolve this limitation of SUBNET-type methods.


\section*{Declaration of generative AI and AI-assisted technologies in the writing process}
During the preparation of this work the authors used GPT-6 Astra pro in
order to generate the multi head implementation and the port-Hamiltonian
variant of DS3M based on an implementation of DS3M with a single head
provided by the authors. Furthermore, a compact Jupyter notebook was
generated using the same tool to evaluate the idea on the benchmark data
based on established evaluation protocols.
The tool was also used to assist with revising the manuscript text and creating the figures.
After using this tool/service, the authors reviewed and edited the content
as needed and take full responsibility for the content of the publication.

\bibliographystyle{abbrvnat}
\bibliography{ifacconf}

\end{document}